\documentclass{article}

\PassOptionsToPackage{numbers,compress}{natbib}
\usepackage[preprint]{neurips_2026}

\usepackage[utf8]{inputenc} 
\usepackage[T1]{fontenc}    
\usepackage{hyperref}       
\usepackage{url}            
\usepackage{booktabs}       
\usepackage{amsmath}
\usepackage{amsfonts}       
\usepackage{nicefrac}       
\usepackage{microtype}      
\usepackage{xcolor}         

\usepackage{graphicx}
\usepackage{capt-of}
\usepackage{rotating}
\usepackage{longtable,array}

\newcommand\be{\begin{equation}}
\newcommand\ba{\begin{eqnarray}}
\newcommand\ee{\end{equation}}
\newcommand\ea{\end{eqnarray}}

\title{Feature Identification via the Empirical NTK}

\author{%
  Jennifer Lin\\
  Principles of Intelligence\\
  \texttt{jylin04@gmail.com} \\
}

\begin{document}

\maketitle

\begin{abstract}
We provide evidence that eigenanalysis of the empirical neural tangent kernel (eNTK) can surface feature directions in trained neural networks. Across three increasingly realistic settings --  a 1-layer MLP trained on modular addition, a 1-layer Transformer trained on modular addition and the pretrained language model Gemma-3-270M -- we show that top eigenspaces of the eNTK align with ground-truth or  interpretable features. In the modular arithmetic examples, top eNTK eigenspaces align with the Fourier features used by the MLP and the Fourier features at seed-dependent frequencies used by the Transformer to implement known ground-truth algorithms. Moreover, the alignment of the relevant subspaces evolves over training, with its first derivative peaking near the onset of grokking. For Gemma-3-270M, we compute top eNTK eigendirections on a dataset of TinyStories context windows and check their alignment with an automatically-generated set of parts-of-speech and other grammatical feature directions. We find that the alignment of eNTK eigendirections with grammar features outperforms a same-budget baseline of PCA on model activations. These results suggest that eNTK eigenanalysis may provide a new handle towards identifying features in trained models for mechanistic interpretability.

\end{abstract}

\section{Introduction}

Mechanistic interpretability aims to reverse-engineer how neural networks process information at inference time \citep{olah2020zoom}. A major open question is to understand how neural networks represent learned features. Operationally, suppose that an input $x$ to a model can either exhibit a particular human-interpretable property (e.g. a certain grammatical construct or sentiment in a natural language context window) 
	or not. Neural networks often seem to learn to detect such properties. For example, they can be probed to classify the presence or absence of such properties after pretraining with higher-than-chance accuracy \citep{alain2016understanding}. We would like to understand how they do so by finding a function of the model's activations and weights 
\be\label{deff}
F(\mbox{model activations/parameters},x) \rightarrow \mathbb{R}
\ee
that detects the presence of the property with high sensitivity and specificity.

It's a priori unclear how to guess a good candidate answer. Previous work has often started from the assumption that individual neural activations \citep{olah2020zoom}, or sparse linear combinations of them \citep{bricken2023towards,cunningham2023sparse} may furnish good candidate features for interpretability. However, it is now believed that such approaches may provide incomplete and non-canonical dictionaries for feature learning \citep{engels2024decomposing,leask2025sparse,paulo2025sparse} (see \citep{sharkey2025open}
for a review). This motivates the exploration of substantially different approaches.

In this work, we propose a new hypothesis: that top eigendirections of the empirical NTK (eNTK) may track learned features in trained MLP and Transformer models. Given a neural network $f_i(x)$ with weights $W_\mu$ and output classes indexed by $i,j,\dots$, the empirical NTK
\be\label{entk}
K_{ij}(x_1, x_2) =  
\sum_\mu \frac{df_i(x_1)}{dW_\mu}\frac{df_j(x_2)}{dW_\mu} 
\ee
is the kernel consisting of two copies of the Jacobian contracted along the parameter-space direction. On a size $N$ dataset and for a particular choice of indices $i$ and $j$, its top eigendirections are length-$N$ vectors that map each dataset point to a number. They thus have the right type \eqref{deff} to be candidate features over the entries of the test set.
\footnote{Additionally, new model inputs $x$ that are not part of the test set used to define an eigendirection $\psi$ can be scored in the sense of \eqref{deff} as $F_\psi(x) = \sum_i \psi(x_i)K(x_i, x)$ summed over test-set points, for $\psi(x_i)$ the number assigned to each test-set point and $K$ the eNTK itself. This converts each eigendirection to a fully general feature in the sense of \eqref{deff}.} 
 We conjecture that for particular choices of how to handle the class indices that we describe below, these eigendirections may track interpretable features in realistic models.

Our main contributions are as follows: 

\begin{itemize}
\item For a 1L MLP trained on modular arithmetic mod $p$ in the grokking regime, where the ground-truth algorithm implemented by the model is known \cite{gromov2023grokking}, a leading set of $4\lfloor \frac p2 \rfloor$ eNTK eigendirections have the same span as the Fourier features used by the first layer of the model in the ground-truth solution. The next set of $4\lfloor \frac p2 \rfloor$ eNTK eigendirections have the same span as the `sum' and `difference' Fourier features used by second layer of the model. The latter eigenspace aligns continuously with the sum modes throughout training, with the first derivative of the overlap reaching a maximum around the onset of the grokking phase transition. 
\item For a 1L Transformer trained on modular arithmetic mod $p$ in the grokking regime, which is known to implement a similar ground-truth algorithm but at a sparse set of random seed-dependent frequencies  \cite{nanda2023progress}, top layerwise eNTK eigenspaces align with the Fourier features at the key frequencies.
\item For the pretrained language model Gemma-3-270M, top eigendirections of the eNTK computed on a  test set of context windows from TinyStories outperform a same-budget baseline of PCA on model activations at recovering independently-specified grammar (e.g. parts-of-speech and morphology) features on the dataset.
\end{itemize}
This paper is organized as follows. In Section \ref{s2}, we explain how we use the eNTK to find features in trained models. In Section \ref{s3} and \ref{s4}, we use the eNTK to find features in a MLP and Transformer, respectively, trained on modular arithmetic. In Section \ref{s5}, we use the eNTK to find grammatical features in Gemma-3-270M. We conclude with a discussion of future directions in Section \ref{s6}.

\vspace{4mm}
\noindent {\bf Related work.} The NTK is famous for being part of an exact solution in deep learning theory when a model's parameter drift throughout training is parametrically small \cite{jacot2018neural, roberts2022principles}. See Section \ref{sA} for a review. A large body of work analyzes the NTK in this context. However, the models that we study in this paper are {\it not} in this ``lazy" regime. 
	Hence, our main hypothesis is a {\it theoretically-motivated empirical conjecture}, which is outside the scope of the theoretical NTK literature.

Some authors, e.g. \cite{Atanasov:2021aa, fort2020deep, mohamadi2023fast, ortiz2021can} have explored whether kernel regression with the eNTK could provide a good approximation to the function learned by a neural network at the end of training outside the lazy regime. (This idea sometimes goes by the name of the ``empirical NTK hypothesis.") The motivation for this idea is similar to the motivation for our work. However, we study not whether the eNTK broadly provides a good approximation to the function learned by a realistic model, but whether natural quantities built from it could be used to interpret features in realistic models.

A few earlier works explored whether eNTK eigenvectors align with interesting directions in trained models. \cite{loo2022evolution} and \cite{tsilivis2022can} visualized kernel eigenvectors  
in small image models (in the context of using the eNTK to study adversarial training), and are the most closely related works to ours. \cite{Atanasov:2021aa} and \cite{baratin2021implicit} demonstrated that eNTK spectral structure can align with target/label directions during training. In contrast, we consider whether eNTK eigendirections can align with latent features.

Besides the eNTK, other objects that involve gradients/curvature have appeared in the mechanistic interpretability literature. As mentioned above, the eNTK is constructed by contracting a pair of Jacobians along the parameter direction, forming a kernel in data space. The Hessian, the second derivative of the loss function w.r.t. model parameters $H = \nabla^2_W \mathcal{L}$, can be approximated by a pair of Jacobians contracted along the opposite (data-space) direction 
	and has appeared in recent interpretability literature. For example, it forms part of the definition of the ``influence function" $\mathcal{I}(z) = \nabla_W f(z')^T H^{-1}\nabla_W \mathcal{L}(z)$, which quantifies the influence of a data point $z$ on a particular test prompt-completion pair $z'$ \cite{koh2017understanding} and was recently used to interpret LLMs at scale \cite{grosse2023studying}. (However, the appearance of the {inverse} Hessian in the influence function {suppresses} leading Jacobian singular directions, so the influence function and top eNTK eigendirections are not directly related at the technical level.) 
	The Hessian also plays a central role in singular learning theory, which has motivated recent interpretability  methods. For example, the ``Local Interaction Basis" algorithm aims to find interpretable directions in a model's activation space in part by aligning would-be directions with singular directions of the layer-to-layer Jacobian \cite{bushnaq2024local, bushnaq2024using}.

\section{Methods}\label{s2}

The main idea that we pursue in this paper is to check if across several models and datasets, top eigendirections of the eNTK \eqref{entk} align with independently-specified features across test set points. Concrete examples of such features  will be presented section-by-section below. One example is whether a language model context window contains a certain grammatical part of speech.

To compute if a length $N$ eNTK eigenvector for $N$ the dataset size aligns with a ground-truth feature, we can take the absolute value of its cosine similarity with a length-$N$ ``feature vector" whose entries consist of that feature's activation on each data point. This gives a scalar score for each (eigenvector, feature) pair. To compute if a size $k$ subspace of eNTK eigendirections aligns with a size $k$ ground-truth feature subspace, we can take the Frobenius norm between the spaces. 

\subsection{Variants of the empirical NTK used in this paper}

The eNTK evaluated over all points in a dataset has shape $(N,N,C,C)$ for $N$ the size of the dataset, and $C$ the number of output neurons (classes). To do an eigendecomposition, we must reduce it to two  dimensions. Below, we'll use two collapses of $K$:

\begin{enumerate}
\item Per-class eNTK. For a 
	class $c \in 1, \dots, C$, we can study the kernel $K_{cc}(x_1, x_2)$.
\item Flattened eNTK. Alternatively, we can form a $NC$ by $NC$ 	matrix by stacking the per-class blocks $\{K_{ij}\}_{i,j =1}^C$ row-major in $i$ and column-major in $j$.
\end{enumerate}

Another variant that we'll use is the layerwise NTK where rather than sum over all model parameters in \eqref{entk}, we sum only over those parameters belonging to a particular layer, before applying either collapse above. This will prove useful for attributing features to a layer.

\subsection{Algorithm for approximating top eNTK eigendirections in large models}\label{s22}
In toy models, we can explicitly materialize the eNTK \eqref{entk} and eigendecompose it with a standard solver. In larger models such as the language model that we study in section \ref{s5}, this is infeasible because the Jacobian used to form the eNTK quickly becomes intractably large. \footnote{In particular, the Jacobian of a size $N$ dataset over $C$ classes contains $N*P*C$ floating-point numbers, for $P$ the number of model parameters. For a $N = 1000$-point test set, a $P=100$M-parameter language model and $C$ = 100 (say) effective classes, this would take 40,000 GB of memory to materialize in fp32.} Instead, we use $2*k$ steps of Lanczos iteration \cite{lanczos1950iteration} to approximate the top $k$ eigendirections of the eNTK. 

The Lanczos algorithm accesses the eNTK only through matrix multiplication on auxiliary vectors $v$, with the computational bottleneck being $2k$ such matrix-vector products. Writing the eNTK as $K = JJ^T$ \eqref{entk}, we can compute
\be
Kv = J(J^T v)
\ee
using a vector-Jacobian followed by a Jacobian-vector product evaluated by autodiff (see e.g. \cite{baydin2018automatic}), without ever materializing either the eNTK or the Jacobian. The products are accumulated over minibatches to fit in GPU memory.

\subsection{Recovering top eigendirections of the flattened eNTK from final hidden-layer gradients}
\label{s23}

In language models, the number of classes $C$ that the model can predict at each sequence position is the vocabulary size $d_{vocab}$. This is often much larger than the size $d_{model}$ of the model's residual stream, and can be intractably large for flattened eNTK computations. On the other hand, the Jacobians at the final hidden layer and output layer are linearly related by the unembedding matrix $U$. 
	Hence, the flattened eNTK $K_{out}$ at the output layer is related to the ``eNTK" $K_r$ built from gradients at the final hidden layer
\be\label{kr}
(K_r)_{ab}(x_1, x_2) = \sum_\mu \frac{dr_a(x_1)}{dW_\mu}\frac{dr_b(x_2)}{dW_\mu}\,,
\ee
where $r_a(x)$ is the final residual layer with indices $a,b \in 1, \dots, d_{model}$ running over the neurons of the layer,
by the relation $K_{out} = (1_N \otimes U) K_r(1_N \otimes U^T)\,$ on a size $N$ dataset. We can use \eqref{kr} to recover the top eigendirections of $K_{out}$ without materializing any intermediate size $d_{vocab}$ objects.

Concretely, given a dataset of size $N$, define the operator $S = I_N \otimes (U^T U)$. Define the isometry 
\be
\Pi = (I_N \otimes U) * S^{-1/2}\,. 
\ee
Then the shape $(N*d_{model}, N*d_{model})$ object
\be\label{ktilde}
\tilde K = S^{1/2} K_r S^{1/2}
\ee
is related to $K_{out}$ by $K_{out} = \Pi\tilde K\Pi^T$; 
has the same eigenvalues as $K_{out}$; and has eigenvectors $q$ related to the eigenvectors $y$ of $K_{out}$ by 
\be\label{evecmap}
y = \Pi q\,.
\ee So we can materialize the top eigenspectrum of $K_{out}$ by running the algorithm of section \ref{s22} on $\tilde K$.

\section{Results for a MLP trained on modular arithmetic}\label{s3}

In our first experiment, we study alignment of top eNTK eigenspaces with the ground-truth Fourier features in a MLP trained on modular arithmetic \cite{gromov2023grokking}.

\subsection{Review of the setup}

In this experiment, for a fixed integer $p$ we construct a dataset consisting of the (length $2p$)  $p^2$ unique pairs of concatenated one-hot encoded integers from $(0, \dots, p-1)$, labeled by their (length $p$) one-hot-encoded sum mod $p$. We split the dataset into non-overlapping train and test sets controlled by a fractional hyperparameter $\alpha = |\mathcal{D}_{\rm train}|/p^2$. We then train a 1L MLP with $2p$ input neurons, $n$ hidden neurons, $p$ output neurons, a quadratic activation function, and no biases to learn the training set with a MSE loss function and AdamW optimizer. Explicitly, the model architecture is
\be\label{mamodel}
f(x) = W^{(2)}(W^{(1)} x)^2\,.
\ee
In the experiments below, we report results for a particular training run at $p = 29$, $n = 512$ and $\alpha = 0.7$. However, we find qualitatively similar results for different seeds and other values of $p$. (See Appendix \ref{robust} for a robustness check.)

With these hyperparameters, the model exhibits grokking \cite{nanda2023progress, power2022grokking}. Some time after reaching 100\% accuracy on the training set, it suddenly goes from 0\% to 100\% accuracy on the test set (see the left panel of Fig. \ref{fig1}), suggesting that it switched to using a more generalizable representation of the underlying modular arithmetic algorithm.

Moreover, in this case we know the exact solution that the model learns after grokking \cite{gromov2023grokking}. At the end of training, the model approximately learns 
known weights s.t.
for input integers $a$ and $b$ the first-layer preactivations have the form $\cos(2\pi \frac k p a)$, $\cos(2\pi \frac k p b)$ up to phases, for $k \in 1, ... \lfloor  \frac p 2 \rfloor$ (below: the 
 $``\cos a"$ and $``\cos b"$ feature families). \footnote{Fourier modes with $k > \lfloor \frac p 2 \rfloor$ are redundant because of trigonometric identities relating arguments $x$ with $2\pi-x$.}
After applying trigonometric identities, the second-layer preactivations turn out to contain sums of terms of the form $\cos(2\pi \frac k p (a+b))$, $\cos 2\pi \frac k p (a-b))$ up to phases, for $k \in 1, ... \lfloor  \frac p 2 \rfloor$ (below: the ``sum" and ``difference" feature families). Most combinations of the second-layer activations cancel on average due to random phases, but one specific combination $\sum_{k=1}^n \cos (2\pi kp (a+b-q))$ survives due to synchronized phases and acts as a modular delta function, enforcing the modular addition solution.

\subsection{Results}

\noindent
\begin{minipage}[t]{0.45\textwidth}
  \vspace{0pt}
  \vspace{8mm}
 {\bf Spectrum.} The per-class eNTK eigenspectrum of the modular-arithmetic model trained to convergence, averaged over classes, is shown in Fig. \ref{maspectrum}. It contains two ``cliffs" or contiguous blocks of eigendirections before and after which there are sharp spectral discontinuities. Each cliff has size 56 = 4$\lfloor \frac p 2 \rfloor$ for our choice of $p = 29$, matching the size of two Fourier feature families. (More precisely, each cosine feature family has size $\lfloor \frac p2 \rfloor$ and comes paired with a sine family for a count of $\{\cos, \sin\} \times \{k = 1, \dots, \lfloor \frac p2 \rfloor\}$ = 2 $\lfloor \frac p2 \rfloor$, so 4$\lfloor \frac p2 \rfloor$ has the right counting to contain two such families.)
\end{minipage}
\hfill
\begin{minipage}[t]{0.45\textwidth}
  \vspace{0pt}
  \centering
  \includegraphics[width=\linewidth]{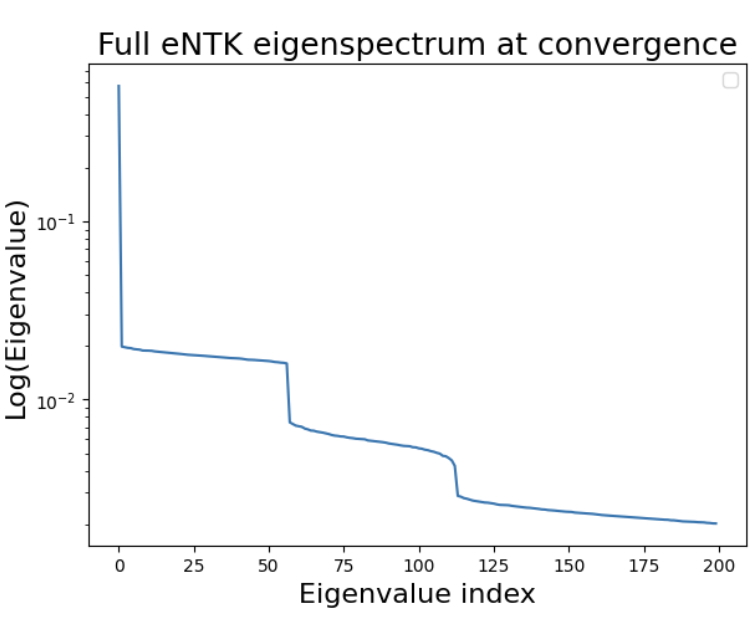}
  \captionof{figure}{Class-averaged eNTK eigenspectrum of the 1L MLP trained on the modular arithmetic task to convergence.}
  \label{maspectrum}
\end{minipage}

\vspace{4mm}
{\bf Top eNTK eigenspaces align with Fourier feature spaces.} Since the cliffs have the same dimensions as pairs of Fourier feature families, it seems plausible that the span of the eigenvectors contained in each of them would overlap with the span of exactly two Fourier families. To test this, we compute the squared Frobenius norm between the cliff directions and each Fourier family, i.e. the sum of squared inner products between all pairs of directions of a normalized basis for the cliff and each feature family. The results, shown in Table \ref{t1}, show that the span of the first cliff contains the $a$ and $b$ Fourier families and the span of the second cliff contains the $a \pm b$ Fourier families.

\begin{table}[h]
  \caption{Squared Frobenius norm of the eigendirections in each cliff with different Fourier feature families, as well as a control of normalized random vectors of the same dimension. A score of 28, which is the dimension of each Fourier feature family, would denote perfect alignment. We find a strong alignment of the first cliff with the $a$ and $b$ Fourier feature families, and of the second cliff with the $a \pm b$ feature families.}
  \label{t1}
  \centering
    \begin{small}
      \begin{sc}
        \begin{tabular}{lccccr}
          \toprule
          Cliff  &  a & b & sum & diff & ctrl  \\
          \midrule
          1    & {\bf 27.54} & {\bf 27.56} & 0.13 & 0.38 & 1.83 \\
          2    & 0.27 & 0.25 & {\bf 26.53} & {\bf 24.86} & 1.80 \\
          \bottomrule
        \end{tabular}
      \end{sc}
    \end{small}
\end{table}

\begin{figure}
\centering
\includegraphics[width=0.88\linewidth]{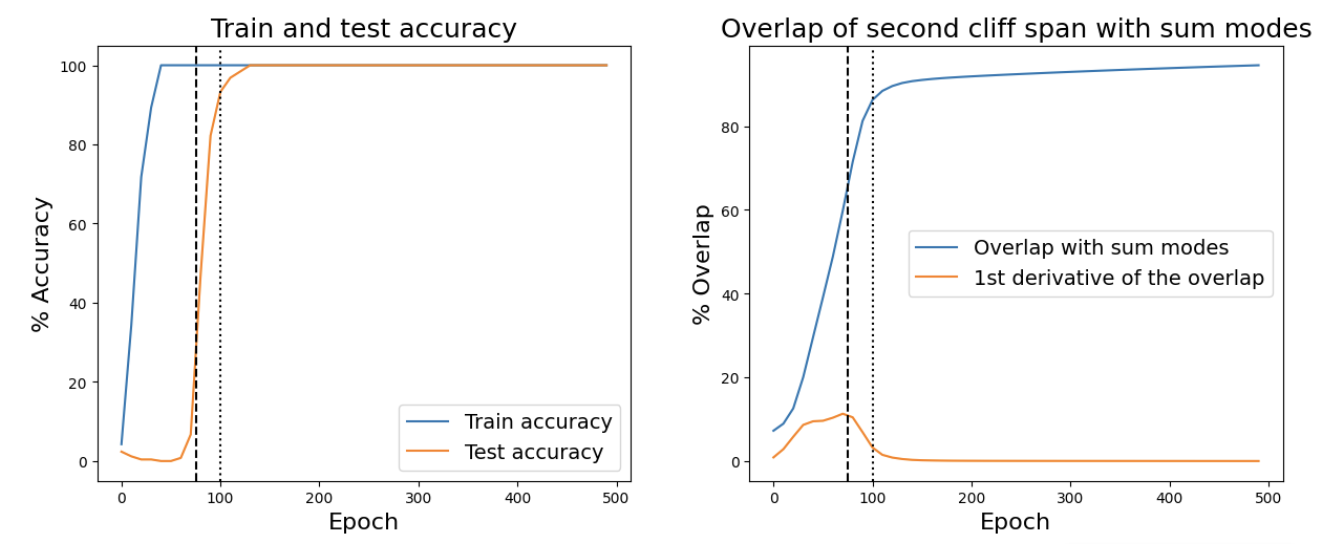}
    \caption{
    Results over time for the MLP trained on modular arithmetic. 
    Left: Train and test accuracy show evidence for a grokking phase transition (sudden onset of generalization) between epochs 75 and 100 (dashed and dotted lines respectively).
   	Right: Alignment of the span of eNTK eigendirections in the ``second cliff" (at eigenvalue indices 57-112) with the sum Fourier modes across checkpoints. The dashed and dotted lines again denote epochs 75 and 100, at the onset and end of the grokking transition.}
    \label{fig1}
\end{figure}

\vspace{4mm}
{\bf Time evolution of the subspace alignment.} The alignment of the eigendirections in the second cliff  
with the sum and difference Fourier modes is small at initialization and rises smoothly  until around the end of the grokking phase transition (see the right panel of Fig. \ref{fig1}). Moreover, the first derivative of the overlap reaches a local maximum around the onset of the grokking phase transition. 

On the other hand, the eigendirections in the first cliff overlap with the $a$ and $b$ modes already at initialization. This follows from the structure of the kernel, as we explain briefly in Appendix \ref{kernels}, and suggests (in combination with the discussion there) that this model's use of Fourier features could perhaps have been predicted at initialization from the eNTK.

\section{Results for a Transformer trained on modular arithmetic}\label{s4}

Next, we study the alignment of eNTK eigenspaces with ground-truth Fourier features in a Transformer trained on modular arithmetic with cross-entropy loss \cite{nanda2023progress}. This model learns a different ground-truth algorithm than the MLP that was the subject of section \ref{s3}.

\subsection{Review of the setup}
In this experiment, for a fixed integer $p$ we construct a dataset consisting of the $p^2$ unique pairs of integers from $(0, \dots, p-1)$ represented as a sequence of three tokens $``[a, b, =]"$, labeled by their one-hot-encoded sum mod $p$ at the final token position. As in the previous experiment, we split the dataset into non-overlapping train and test sets controlled by a fractional hyperparameter $\alpha$. We then train a 1-layer GPT-type Transformer, without LayerNorm and with untied embedding/unembedding matrices, to learn the training set with cross-entropy loss and an AdamW optimizer.  For a review of the exact architecture, see Appendix A of \cite{nanda2023progress}.

Below, we report results for $p = 29$, $d_{model} = 64$, 4 attention heads of dimension $d/4 = 16$, and $n=256$ hidden units in the MLP. We use learning rate 3e-4, weight decay parameter 1.0, and $\alpha = 0.5$. With these hyperparameters, the model exhibits grokking across several seeds.

The ground-truth modular arithmetic algorithm learned by this model after grokking is different from the one learned by the MLP in section \ref{s3} \cite{nanda2023progress}. One major difference is that rather than learn Fourier modes at all intermediate frequencies, the model learns only Fourier modes at a sparse set of random, seed-dependent frequencies. Specifically, at the embedding step, the model maps the inputs $a$ and $b$ to sines and cosines at a set of key random frequencies $\omega_k = 2\pi k/p$, $k \in \mathbb{N}$.  
After the embedding step isolates the key frequencies, the model then computes the trigonometric sum identities for $\cos(\omega_k(a+b))$ and $\sin(\omega_k(a+b))$ on those frequencies only in the attention and MLP layers, and the trigonometric difference identity $\cos(\omega_k(a+b-c))$ on those frequencies only in the MLP output and unembedding layers. Finally, it uses the unembedding matrix to add together the $\cos(\omega_k(a+b-c))$'s for different $k$'s. This causes destructive interference unless $a+b-c=0$.

Because we do not use LayerNorm in the architecture (in keeping with the conventions of \cite{nanda2023progress}), the eNTK spectra at different layers of the model differ from each other by orders of magnitude, and the full spectrum is dominated by the largest such contributions. It's therefore more meaningful to look at the layerwise eNTK spectra. 

Because the key frequencies in this experiment are seed-dependent, we will analyze a particular training run in this section which had two key frequencies at $k = 1$ and $k=12$. However, we checked that results at other seeds are qualitatively similar in appendix \ref{robust}.

\subsection{Results}

\begin{figure}
\centering
\includegraphics[width=0.88\linewidth]{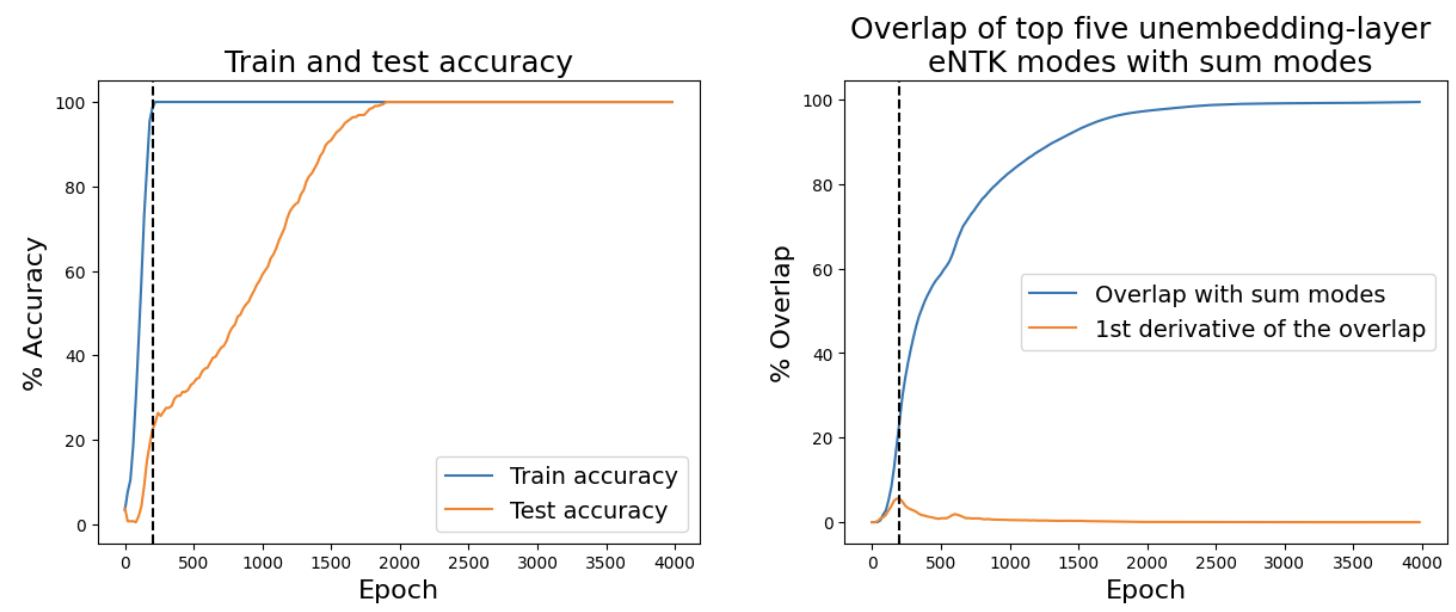}
    \caption{Results over time for the Transformer trained on modular arithmetic. Left: Train and test accuracy show evidence for a grokking phase transition beginning around epoch 200 (dashed line). Right: Alignment of the top five eigendirections of the eNTK at the unembedding layer with the sum Fourier modes across checkpoints. The dashed line again denotes epoch 200.
    }
    \label{fig2}
\end{figure}

\noindent {\bf Top eNTK eigenspaces align with Fourier features at key frequencies.} In this training run, the leading five eigendirections of the layerwise eNTK associated with the O+V parameters in the attention block, the input matrix $W_{in}$ of the MLP layer, and the output matrix $W_{out}$ of the MLP layer each overlap significantly with the sum of $a$ and $b$ Fourier features $\{\cos(\omega_ka) + \cos (\omega_kb), \sin (\omega_ka) + \sin (\omega_kb)\}$ at the key frequencies, while the top five eigendirections of the eNTK for the unembedding matrix overlap significantly with the `sum' Fourier features. To see this, we compute the squared Frobenius norm between said eNTK eigendirections and the corresponding feature families at the key frequencies in Table \ref{t2}. The number five is one more than twice the number of key frequencies in this run, suggesting that the eigendirections mix precisely those Fourier features with a zero mode.
We also find that the next four eigendirections of the $W_{out}$ and unembedding layers overlap with the `sum' and $a$ and $b$ Fourier feature families, respectively.

\begin{table}
  \caption{Squared Frobenius norm of the eigendirections mentioned in the text with different Fourier feature families at the key frequencies, as well as a control of normalized random vectors of the same dimension. A score of 4 would denote perfect overlap with the Fourier feature directions.}
  \label{t2}
  \begin{center}
    \begin{small}
      \begin{sc}
        \begin{tabular}{lccccr}
          \toprule
          layer  & A+B & sum & diff & ctrl  \\
          \midrule
          OV    & {\bf 3.86} & {0.12} & 0.04 & 0.16 \\  
          MLP$_{in}$   & {\bf 3.89} & 0.05 & 0.01  & 0.14 \\ 
          MLP$_{out}$ (1)   & {\bf 3.90} & 0.02 & 0.03 & 0.14 \\ 
          MLP$_{out}$ (2)   & 0.02 & {\bf 3.94} & 0.01 & 0.14 \\
          U (1)   & 0.02& {\bf 3.97} & 0.02 & 0.14 \\
          U (2)   & {\bf 3.92} & 0.00 & 0.02 & 0.10 \\ 
          \bottomrule
        \end{tabular}
      \end{sc}
    \end{small}
  \end{center}
\end{table}

\vspace{4mm}
\noindent {\bf Time evolution of the subspace alignment.} For the first set of unembedding-layer eNTK eigendirections that align with the Fourier sum modes at convergence, we checked that the alignment of the eigendirections with the sum modes rises smoothly throughout the grokking transition and the first derivative of the overlap reaches its maximum around the onset of the grokking phase transition, in agreement with the similar analysis in section \ref{s3}.

\section{Finding grammar features with the eNTK in Gemma-3-270M}\label{s5}
Finally, we compute top eNTK eigendirections in the Gemma-3-270M language model over a small test dataset and evaluate their alignment with independently-generated grammar feature directions.

\subsection{Setup}
In this experiment, we start with the Gemma-3-270M language model \cite{2025arXiv250319786G}: a pretrained and instruction-tuned GPT-type model with 18 Transformer blocks, each with $d_{model} = 640$, and a tokenizer with a vocabulary size of 262144. 

We construct a stratified dataset consisting of $N=1024$ length-64 context windows from TinyStories \cite{eldan2023tinystories} as follows. For each of eight grammatical parts of speech (POS's), we collect 128 length-64 random context windows from across TinyStories where both the next word in  TinyStories itself and the next word predicted by the model has that part of speech, as tagged by the Stanza NLP toolkit \cite{qi2020stanza}. We also keep track of additional linguistic features for the model-predicted next word, as tagged by Stanza: for example whether a verb is in past tense. 
	Operationally, we retain all Stanza-labeled morphological features appearing above a threshold number of ($k=30$) examples and ($d=12$) distinct words, excluding labels highly redundant with a POS tag or with another retained feature.
	The eight parts of speech and additional linguistic features provide a set of automatically-generated features of the language model that we can look for with the eNTK eigendirections.
See Appendix \ref{data} for details about the dataset. See Appendix \ref{robust} for a robustness check on the results reported here with different random draws of the dataset.

Using the algorithm described in Section \ref{s22}, we approximately compute the top 25 eigendirections of the layerwise object $\tilde K$ \eqref{ktilde} whose eigendirections are linearly related to those of the full flattened eNTK as explained in Section \ref{s23},  at each of the 18 model layers w.r.t. the parameters in the MLP$_{out}$ parameter matrix at that layer. We focus on the MLP$_{out}$ parameters as a computationally tractable layer that directly writes transformed features to the residual stream.

We empirically observe that for all shape $(N, d_{model})$ eigenvectors thus computed, the majority of the linear projection along singular directions lies along the top singular direction. Hence, for each eigenvector we assign a scalar score to each dataset point by scoring examples by the top left singular vector. I.e. for each eigenvector, we compute the SVD $U\Sigma V^T$, then take the score of example $i$ under mode $m$ to be $u_{m,1}(i)\sigma_{m,1}$. Since the map \eqref{evecmap} from each $\tilde K$ eigenvector to an eigenvector of the output-layer eNTK is an isometry on the residual/vocabulary coordinate, the scores thus obtained would be the same as those for the corresponding output-layer eNTK up to sign.

Finally, for each part of speech and additional linguistic feature tracked in the dataset, we find the eNTK eigendirection among the ones computed that best aligns with the feature by AUROC score, with the following protocol. We randomly split the dataset into train/test halves; for each binary feature, choose the signed mode that best aligns with it on the training set; and report the AUROC score on the test set.

\subsection{Results}

\begin{table}
  \caption{AUROC score of the best eNTK eigendirection and the best PCA direction on model activations from among the top 25 such directions computed at each model layer, on each grammar feature tracked in the dataset, as evaluated with the protocol described in the main text. Also shown is the model layer that the best eNTK eigenvector comes from.}
  \label{t3}
  \begin{center}
    \begin{small}
        \begin{tabular}{lcccc}
          \toprule
           Feature & Feature type & Best eNTK AUROC & Best PCA AUROC & best eNTK layer \\
          \midrule
          ADV & pos & {\bf 0.838274} & 0.789272 & 15 \\
          CCONJ & pos & {\bf 0.850237} & 0.797049 & 13 \\
          DET & pos & {\bf 0.758161} & 0.717041 & 16 \\
          NOUN & pos & {\bf 0.993652} & 0.964007 & 15 \\
          PRON & pos & {\bf 0.802595} & 0.751779 & 13 \\
          PROPN & pos & {\bf 0.943080} & 0.921666 & 12 \\
          PUNCT & pos & {\bf 0.888951} & 0.863630 & 13 \\
          VERB & pos & {\bf 0.968436} & 0.934152 & 15 \\
          Number=Plur & morph & {\bf 0.798755} & 0.718326 & 14 \\
          Number=Sing & morph & {\bf 0.805700} & 0.732243 & 16 \\
          Person=3 & morph & 0.841222 & {\bf 0.858841} & 15 \\
          Tense=Past & morph & {\bf 0.957330} & 0.950598 & 16 \\
          VerbForm=Fin & morph & {\bf 0.987603} & 0.979670 & 14 \\
          VerbForm=Inf & morph & {\bf 0.999525} & 0.983482 & 16\\
          \bottomrule
        \end{tabular}
    \end{small}
  \end{center}
\end{table}

{\bf eNTK eigendirections align with grammar features better than a PCA baseline.} Our main result in this section is that the layerwise eNTK eigendirections recover the POS and morphology labels on the dataset better than a same-budget PCA baseline. 
Namely, at each model layer we also compute the top 25 PCA directions on final residual stream activations at that layer. For each part of speech and additional linguistic feature that we  track in the dataset, we find the PCA direction that best aligns with the feature by AUROC score with the same protocol. The head-to-head results are summarized in Table \ref{t3}, where eNTK eigendirections outperform PCA of model activations on all POS features and all but one morphological feature by AUROC score. Both substantially outperform the random baseline of 0.5 across all categories.

We choose PCA on model activations as the baseline because it provides a same-data baseline on the small test set. In contrast, a $65k$-token dataset is likely insufficient to produce a training signal for a SAE. We consider the ability of eNTK eigenanalysis to surface feature directions with small test sets to be one of its selling points.

\vspace{4mm}
{\bf Qualitative analysis.} In Appendix \ref{lmex}, we show examples of select eNTK eigendirections whose top-activating dataset examples suggest  specific qualitative interpretations.

\section{Discussion} \label{s6}

To summarize, we have provided evidence that eigenanalysis of the eNTK can surface learned features across three increasingly realistic settings where the features can be independently specified: a 1L MLP trained on modular arithmetic and a 1L Transformer trained on modular arithmetic, where top eNTK eigenspaces recover the Fourier feature subspaces known to be used by the learned algorithms, and Gemma-3-270M, where top eNTK eigendirections on a test set recover grammatical features better than a same-budget PCA baseline on model activations.
Together, these results suggest that eNTK eigenanalysis may provide a useful new handle towards identifying features in trained models for mechanistic interpretability.

\vspace{4mm}
{\bf Limitations and future work.} The limitations of this work suggest avenues for future exploration. The language model experiment presented here, while suggestive, is nonetheless limited in several ways: it takes place on a smaller-than-realistic model (Gemma-3-270M), with a very small (1024-point) test set, drawn from an especially simple data distribution (TinyStories). It would be interesting to generalize our analysis to larger models and datasets. It would also be interesting to study the eigendirections of the eNTK across multiple parameter matrices and perhaps multiple layers at a time, instead of only the per-layer MLP$_{out}$ weight matrices considered in the language model experiment. One could also study the eNTK formed w.r.t. gradients for more than next token prediction. 

The results presented in this paper are empirical.
It would be interesting to understand analytically why the eNTK recovers the observed feature directions, perhaps starting with the modular arithmetic example where the signals are especially clean. As a starting point for progress in this direction,  Appendix \ref{kernels} gives closed-form expressions for the relevant kernels and explains why the  first eigenvalue ``cliff" in the modular arithmetic experiment appears already at initialization.

Another limitation is that the results presented in this paper are correlational, not causal. Although we find that eNTK eigendirections align with independently-specified features, we have not shown that an eNTK-inspired intervention would change the model behavior. 
Progress on this direction would be helpful for potential practical applications.

Finally, we believe that the results on time evolution in the modular arithmetic experiments suggest a potential use of eNTK eigenanalysis beyond static feature discovery. There, we saw that the alignment of eNTK eigendirections with the relevant Fourier feature directions grows rapidly near the grokking transition. If analogous signals could be identified in less controlled settings, perhaps eNTK subspace alignment could provide a way to monitor when particular features are acquired during training, or to identify which training examples contribute to the feature formation. 
	We believe that future work on this direction could be an interesting next step with an eye towards developing practical diagnostics for AI safety applications.

\newpage

\begin{ack}
I am grateful to Ari Brill, Tom Ingebretsen Carlson, Lauren Greenspan, Andrew Mack, Nischal Mainali, Logan Smith, Lucas Teixeira and Dmitry Vaintrob for discussions and comments on the manuscript. I am especially grateful to Andrew Mack for several conversations about scaling to language models. This work was supported by PIBBSS and the Long-Term Future Fund.
%
%
\end{ack}

{\small
\bibliographystyle{plainnat}
\bibliography{mlp_ntk}
}

%
%
%
%
%
%


\appendix

\section{Review of the theoretical NTK} \label{sA}

In this section, we briefly review some key results from the theory 
of the neural tangent kernel (NTK) \eqref{entk}, which provided initial motivation for our hypothesis in the main text. 

The NTK theory \cite{jacot2018neural, roberts2022principles} gives a closed-form formula for the function $f$ learned by a neural network under gradient descent with MSE loss in a ``lazy limit" where the drift in model parameters throughout training is parametrically small. In this regime, the model provably learns the function  
 \be\label{linearized_approx}
f_i(x) = \sum_{\alpha_1, \alpha_2 } K_{ij}(x, x_{\alpha_1}) K^{-1}_{jk}(x_{\alpha_1}, x_{\alpha_2})\, y_{k, \alpha_2}\,,
 \ee
 where the indices $i,j,k$ run over the output neurons (classes), the $\alpha$'s run over the points in the training set, and $K_{ij}$ is the NTK, \eqref{entk}.

The idea behind the derivation is that if the changes in a model's weights $W_\mu$ during training are small enough that we can truncate a Taylor expansion of the model in its weights to first order while maintaining a good approximation, then perhaps we can resum the {linearized approximation} to how the model changes at each step of gradient descent to get a closed-form formula for the function learned by the model. 

In equations, this idea plays out as follows. Given a neural network $f_i(x, W_\mu)$ with index $i$ running over the output neurons, the Taylor expansion for the change in the model after a step of gradient descent is

\be\label{e3}
f_i(x, W_\mu(t+1)) - f_i(x, W_\mu(t)) = \left. \sum_\mu \frac{df_i(x)}{dW_\mu}\right|_{W_\mu = W_\mu(t)} dW_\mu + \dots
\ee

where the $\dots$'s denote terms quadratic and higher in $dW_\mu$. Using the definition of gradient descent to replace $dW_\mu \rightarrow -\eta \frac{d\mathcal{L}}{dW_\mu}$, for $\eta$ the learning rate and $\mathcal{L}$ the loss function,  as well as the fact that the loss function itself depends on the model evaluated on  points in the training set, 
we can massage \eqref{e3} into 
\be\label{e5}
f_i(x, W_\mu(t+1)) - f_i(x, W_\mu(t)) = \left.-\eta \sum_j \sum_{\alpha\in \mathcal{D}} \frac{d\mathcal{L}}{df_j(x_\alpha)}\left[\sum_\mu \frac{df_i(x)}{dW_\mu}\frac{df_j(x_\alpha)}{dW_\mu} \right]\right|_{W_\mu = W_\mu(t)} + \dots \,.
\ee
The term in brackets on the RHS is the NTK, \eqref{entk}. \footnote{Throughout this section we also use that the NTK itself is time independent in the lazy limit.}

Now suppose that we're {allowed} to truncate the $+ \dots$ while maintaining  a controlled approximation to the function learned by the neural network. Moreover, let's specialize to MSE loss. This yields 
\be\label{e6}
f_i(x, W_\mu(t+1)) - f_i(x, W_\mu(t)) = -\eta\sum_j \sum_{\alpha\in \mathcal{D}}(f_j(x_\alpha, t) - y_{j, \alpha}) K_{ij}(x, x_\alpha)\,.
\ee 
If we could sum the LHS of \eqref{e6} over all time steps $t$, we would get the desired closed-form formula for the function learned by the neural network in the linearized approximation. What remains is to get  rid of the $t$-dependence on the RHS of \eqref{e6}. To do this, we apply \eqref{e6} itself to the special case that the argument $x$ on the LHS is a point $x_\alpha$ in the training set, yielding roughly $f_j(x_\alpha, t) - y_{j,\alpha} = (1 - \eta K)^{t+1}(f_j(x_\alpha, t=0) - y_{j, \alpha})$ up to indices. This turns the sum over time steps into a geometric series in $(1 - \eta K)$ whose resummation yields Eq. \eqref{linearized_approx}.

The mathematical assumption required for these steps to go through is that we must be able to discard the higher-order terms in eqs. \eqref{e3} and \eqref{e5}. Roughly, this assumption can be imposed at the level of the model architecture 
 in a scaling limit where we make the model much wider than it is deep, while initializing the weights with a 1/$\sqrt{\mbox{width}}$ scaling. See \cite{roberts2022principles} for a recent pedagogical review.

However, realistic models are usually not in this regime. In particular, we emphasize that the models studied in the main text are not in the lazy regime. Hence, our study is a {\it theoretically-motivated empirical} study. Since on the one hand exact solutions to deep learning theory anywhere in parameter space are few and far between, and on the other hand we lack a principled way to pick candidate features for mechanistic interpretability, we simply speculated that we could extrapolate the NTK far beyond its regime of validity and use natural quantities built from it to interpret realistic models. 
The surprising result is that this idea appears to work in the situations considered in the main text.

\section{Kernels used in this paper} \label{kernels}

\subsection{Kernels for the MLP/modular arithmetic experiment}

For the MLP/modular arithmetic experiment, the model architecture $f(x) = W^{(2)}(W^{(1)}x)^2$ \eqref{mamodel} implies the layer-1 Jacobian 
\be
\frac{\partial f_i(x)}{\partial W^{(1)}_{km}} = 2 W^{(2)}_{ki} (W^{(1)} x)_k x_m
\ee
and hence, the layer-1 kernel
\be\label{l1k}
K_{ij}^{(1)}(x_1, x_2) = 4 \sum_k W^{(2)}_{ki} W^{(2)}_{kj} (W^{(1)}x_1)_k(W^{(1)}x_2)_k (x_1 \cdot x_2)\,.
\ee
The layer-2 Jacobian is
\be
\frac{\partial  f_i(x)}{\partial W_{ki'}^{(2)}} = \delta_{ii'}(W^{(1)}x)^2_k, 
\ee
hence the layer-2 kernel is
\be
K_{ij}^{(2)}(x_1, x_2) = \delta_{ij} \sum_k (W^{(1)} x_1)_k^2 (W^{(1)}x_2)_k^2\,.
\ee
The total kernel is their sum, $K = K^{(1)} + K^{(2)}$.

It would be interesting to understand which properties of the weights and data lead to the full mechanistic structure described in the main text, including the time-evolution results shown in Fig. \ref{fig1}. In general, we will leave this for future work. 

As a warm-up however, note that the appearance of the first eigenvalue cliff at initialization can be explained by the structure of Eq. \eqref{l1k}. Eq. \eqref{l1k} contains a parameter-agnostic factor of $x_1 \cdot x_2$ which is zero for one-hot-encoded data points $(a,b)$ and $(a', b')$ unless either $a = a'$ or $b=b'$. This makes the layer-1 kernel have nonzero values along each fixed-$a$ and fixed-$b$ line, and zeroes elsewhere.  Moreover, the nonzero kernel entries at initialization are approximately identical for each choice of $a$ or $b$ by symmetry. Hence, the kernel at initialization has an approximate shift symmetry along the $a$ and $b$ axes, so is approximately diagonalized by Fourier waves on each axis, with (nonzero) eigenvalues related to the weight initializations.  

	In this example, the $x_1 \cdot x_2$ factor in \eqref{l1k} relied on the model architecture (specifically, the fact that each one-hot-encoded data point only touches two columns of the layer-1 weight matrix), and its conversion to a shift-symmetric kernel relied on the shift symmetry of the dataset. This illustrates a broader point that the NTK (and its eigenstructure) inherits information about both already at initialization. 

\subsection{Kernels for Transformer experiments}

For the 1L Transformer/modular arithmetic experiment in section \ref{s4}, layerwise eNTK formulas for the late layers of the model are as follows (with $i$ a class index $\in$ $d_{vocab}$, $a,b$ indices $\in$ $d_{model}$, and $m,n$ indices $\in$ $d_{mlp}$):

\vspace{2mm}
(*) For the unembedding matrix (of shape ($d_{vocab}$, $d_{model}$)), the per-class layerwise eNTK is a diagonal delta function in class indices $\times$ the data-data Gram matrix of the residual stream at the final sequence ('=') position before unembedding, i.e., the shape $(N,N)$ tensor whose $ij$th entry is the inner product of the residual stream at the `=' position when we evaluate the model on the $i$th point in the dataset and the residual stream at the `=' position when we evaluate it on the $j$th point in the dataset.

	To see this, let $r(x)$ be the residual stream at the final sequence position before unembedding. The model output is 
	\be
	f_i(x) = (W_U)_{ia} r_a(x)\,,
	\ee
	hence 
	\be
	\frac{df_i(x)}{d(W_U)_{ja}} = \delta_{ij} r_a(x)\,.
	\ee
	Hence,
	\be\label{ku}
	K^{W_U}_{ii'}(x, x') = \sum_{j,a} \frac{df_i(x)}{d(W_U)_{ja}}\frac{df_{i'}(x')}{d(W_U)_{ja}} = \delta_{ii'}\sum_a r_a(x) r_a(x')\,.
	\ee

\vspace{2mm}
(*) For the output matrix $W_{out}$ of the MLP layer (of shape $(d_{model}, d_{mlp}))$, the layerwise eNTK is the data-data Gram matrix of the MLP hidden-layer activation at the final sequence (`=') position $\times$ a fixed class-dependent rescaling from $W_U$. 
	
	To see this, note that the MLP block contributes $(W_{out})_{am}\phi(W_{in} r)_m$ additively to the  residual stream, which gets multiplied by $(W_U)_{ia}$ to contribute to the final model output, $f_i(x)$. (The model output $f_i(x)$ also contains earlier contributions to the residual stream, but those don't depend on $W_{out}$.) Hence, 
	\be
	\frac{df_i(x)}{d(W_{out})_{am}} = (W_U)_{ia}\phi(W_{in}r(x))_m\,,
	\ee
	and 
	\be\label{kout}
	K_{ii'}^{W_{out}}(x, x') = \sum_{a,m}\frac{df_i(x)}{d(W_{out})_{am}}\frac{df_{i'}(x')}{d(W_{out})_{am}} = (W_UW_U^T)_{ii'}\sum_a\phi_a(x)\phi_a(x')\,.
	\ee

\vspace{2mm}
(*)  For the input matrix $W_{in}$ of the MLP layer (of shape ($d_{mlp}, d_{model}))$, the layerwise eNTK is the data-data Gram matrix of the residual stream going into the MLP layer at the final sequence (`=') position $\times$ a data-dependent prefactor from the part of the model ``after" $W_{in}$ in the inference-time direction of the model.

	To see this, we again take the starting point that the MLP block contributes $(W_U)_{ib} (W_{out})_{bn}\phi((W_{in})_{nc}r_c)$ to the model output $f_i(x)$.  Hence, 
	\be
	\frac{df_i(x)}{d(W_{in})_{ma}} = (W_U * W_{out})_{im}\phi'_m(x) r_a(x)\,.
	\ee
	Hence, 
	\begin{eqnarray}\label{kin}
	K_{ii'}^{W_{in}}(x, x') &=& \sum_{a,m} \frac{df_i(x)}{d(W_{in})_{ma}}\frac{df_{i'}(x')}{d(W_{in})_{ma}} \\ &=& \sum_m(W_U W_{out} W_{out}^T W_U^T)_{ii'}\phi'(W_{in}r(x))_m\phi'(W_{in}r(x'))_m \sum_a r_a(x)r_a(x')\,.
	\end{eqnarray}
	If the activation function is a ReLU, the data-dependence of this prefactor is just a binary gate that sets entries in $K_{ii'}^{W_{in}}$ to 0 if the arguments are negative.

Note that across the examples \eqref{ku}, \eqref{kout}, \eqref{kin}, the layerwise eNTK factorizes into a kernel built from Jacobians of the model wrt the model activation after the layer $\times$ a data-data Gram matrix of the model activations preceding the layer. In fact, this pattern generalizes to deeper Transformers. In a model with 2+ Transformer blocks, the dependence of the output on a given parameter matrix $W_c$ takes the schematic form 
\be\label{kschem}
f_i = W_U* ((\mbox{current block after $W_c$}) \circ W_c * r_{in}) + W_U * F((\mbox{current block after $W_c$}) \circ W_c * r_{in})\,,
\ee
where $r_{in}$ parametrizes the activations preceding $W_c$. (Here we've suppressed the index structure, but $W_U$ carries the class index $i$ and all others are contracted.) The first term on the RHS of \eqref{kschem} comes from the direct contribution of the current block to the residual stream, and the second, which is absent in eqs. \eqref{ku} - \eqref{kin}, comes from how later blocks use the output of the current block as inputs. Differentiating, we get schematically that
\be\label{dschem}
\frac{df_i(x)}{dW_c} = \left\{W_U* \left(1 + \frac{dF}{d(\mbox{current block})} \right)* \frac{d\mbox{(rest of current block)}}{dW_c}\right\} * r_{in}(x)\,.
\ee
If we ``square" \eqref{dschem} to form the layerwise eNTK, we will find again the same pattern with the bracketed term giving rise to a kernel built from per-class Jacobians of the model wrt the model activation right after $W_c$, along with a class-independent data-data Gram matrix of the model activation right before $W_c$.

\section{Dataset used in the language model experiment} \label{data}

In Table \ref{tgrammar}, we describe the grammatical features used in the experiment in section \ref{s5}. 

\begin{table}
  \caption{Description of automatically-generated grammatical features used in the experiment in section \ref{s5}.}
  \label{tgrammar}
  \begin{center}
    \begin{small}
        \begin{tabular}{llll}
          \toprule
           Feature & Feature name  & Description & examples \\
          \midrule
          ADV & Adverb & Modifies a verb, adjective, adverb or clause. & quickly, very, soon\\
          CCONJ & Conjunction & Connects words, phrases or clauses. & and, or, but\\
          DET & Determiner & Introduces or specifies a noun. & the, this, my\\
          NOUN & Noun & A person/place/thing/... excl. proper names & dog, city, idea \\
          PRON & Pronoun & Substitutes for a noun phrase & I, he, it \\
          PROPN & Proper noun & Proper name of a specific person/place/etc. & Lily, Tim, Joe  \\
          PUNCT & Punctuation & Punctuation marks and separators. & ., !, "\\
          VERB &  Verb & Expresses an action, event or state & run, eat, think \\
          Number=Plur & Plural number &  The word refers to more than one entity. & dogs, children, we \\
          Number=Sing & Singular number & The word refers to one entity. & dog, child, I \\
          Person=3 & Third person & The word refers to someone who is neither & he, she, they   \\ && a speaker or addressee. & (*also verbs referring \\ &&& to the 3rd person) \\
          Tense=Past & Past tense verb & The word marks an action/state in the past & walked, went, ran   \\
          VerbForm=Fin & Finite verb & A verb tied to a subject, that can show tense & walks, went, runs \\
          VerbForm=Inf & Infinitive verb & Base verb form; invariant with subject/tense & to go, to eat, will sit\\
          \bottomrule
        \end{tabular}
    \end{small}
  \end{center}
\end{table}

The eight parts-of-speech chosen as input to construct the stratified dataset were common parts-of-speech appearing across words in the TinyStories dataset. After they were used to generate the dataset, the morphological features were automatically picked by the following rule. We used Stanza to automatically generate morphological labels for all words in the dataset, and retained those labels appearing above the threshold numbers of $k=30$ distinct dataset examples and $d=12$ distinct words. We also excluded labels highly redundant with a part-of-speech tag or with another retained feature. 

To quantify redundancy with a POS tag, for each morphological label $m$ we found the POS tag that co-occurred with it the most, computed purity [ = fraction of examples with label $m$ that had the dominant POS tag] and coverage [ = fraction of examples with the dominant POS tag that had the  label $m$], and excluded morphological tags with both purity and coverage > 0.9.
To quantify redundance among features, we required that the Jaccard overlap $(a \cap b) / (a \cup b)$ between categories not exceed 0.9.

\section{Examples of interpretable eNTK eigenmodes in the language model experiment } \label{lmex}

In this appendix, we show examples of eNTK eigendirections that have especially interpretable top-activating dataset examples according to the signed score defined in section \ref{s5}. For each eigendirection we show the top 20 dataset examples. For each example we show the next word from the TinyStories dataset for that context window, and its part-of-speech. 

Examination of top-activating dataset examples provides an in-principle unsupervised way to assign qualitative semantic meanings to candidate feature directions. For example, such qualitative analysis constituted a large part of the initial evidence for sparse autoencoders as an interpretability tool in \cite{bricken2023towards}.

\subsection{Infinitive verb mode in layer 16} 

\begin{longtable}{@{}r>{\raggedright\arraybackslash}p{0.62\textwidth}>{\raggedright\arraybackslash}p{0.12\textwidth}l@{}}
\hline
\textbf{index} & \textbf{window} & \textbf{Next TS word} & \textbf{POS} \\
\hline
\endfirsthead
\hline
\textbf{index} & \textbf{window} & \textbf{Next TS word} & \textbf{POS} \\
\hline
\endhead
0 & Max tried to run away, but the big dog chased him. Max ran and ran until he got lost.\par Max couldn't find his way back home and he was scared. He looked for a new shelter, but he couldn't find one. As the night got darker and colder, Max started to & look & VERB \\
1 & One day, her mommy asked her to clean up her toys. Lily didn't want to stop playing, so she asked her mommy if she could clean up later. Her mommy said no, and Lily got upset.\par Lily's mommy turned on some playful music to make cleaning up more fun. Lily started to & make & VERB \\
2 & help her, so he walked slowly with her. The little girl was happy and said, "Thank you, Max!"\par Later that day, Max went to the lab with his owner. The lab had lots of fun toys for Max to play with. Max had so much fun, he didn't want to & play & VERB \\
3 & it lit up the whole night sky. It was so beautiful!\par The pot had been sitting in a field of icy snow for many years. Over time, it had grown colder and colder until it looked like a big, icy rock.\par Suddenly, a little spark jumped out of the pot and the copper started to & melt & VERB \\
4 & anything like it before and wanted to learn more.\par She tried to pull it but it was stuck. Lily kept tugging on it until it came undone! She was so excited and wanted to see what was inside. But then she heard her mother calling her to stop. Reluctantly, Lily stopped and went to & sleep & VERB \\
5 & was very hungry and wanted to eat his meal. His mom cooked him some yummy mac and cheese. Timmy was happy and said, "Thank you, mom. This meal is delicious!"\par After he finished eating, Timmy went outside to play. He saw a little bird who was lost and didn't & know & VERB \\
6 & nice and cool. From the top, he could see all of the nice things which happened down below.\par The bear was very friendly. He liked to meet other animals and play games. One day, he saw two dogs who were quarreling. They were fighting over a stick.\par The friendly bear decided to & play & VERB \\
7 & brothers. They like to play with their toys. They have many vehicles: cars, trucks, planes and trains. They make noises and move them around the floor.\par One day, Mom says, "We have to go to the store. Put your toys away and get ready." Tim and Sam do not want to & go & VERB \\
8 & child wearing a bright red coat ran up to the tree. He was so happy as he hugged the tree. He wrapped his arms around the frozen pine and felt its icy needles. They felt like tiny tiny hugs.\par The little boy stayed for a while and the pine tree stayed frozen. But then the sunshine started to & shine & VERB \\
9 & sun on her face and the people around her talking and laughing.\par As the day went on, more and more people joined Maisy at the famous bench. Everyone who came to the park wanted to know why Maisy had chosen to remain there.\par But Maisy just smiled and said nothing. She eventually had to & have & VERB \\
10 & bright and colourful. She ran out to the garden and the kite flew high in the sky. Sally was so happy, she couldn't stop smiling.\par Suddenly the wind picked up, and the kite flew even higher. The strings were pulled tight, but Sally held on tight. She did not want the kite to & fly & VERB \\
11 & !" So, Lily climbed up the ladder and slid down the slide. She loved it! She went down the slide over and over again.\par While Lily was playing on the slide, she saw a boy who was being rude to his friends. He was not being nice and was calling them names. Lily didn't & like & VERB \\
12 & , the animals were ready to play. The chickens would run around and the cows grazed in the grass.\par One day, a cute, little fox came to the ranch. He was so curious about the animals, he wanted to see what they were doing. But the animals were so busy, they didn't even & want & VERB \\
13 & Dad's hands so they would not get lost. They followed the signs to the big building where they had to check in their bags and show their tickets.\par Lily and Ben did not understand why they had to wait in a long line and put their bags on a belt. They did not understand why they had to & wait & VERB \\
14 & they discovered that it was a whale.The whale was so big! They had never seen a whale so large.\par They stopped their boat and watched the whale for a while. It was an amazing sight. They saw the whale diving and swimming around.\par After some time, the friends decided it was time to & go & VERB \\
15 & family and was having the best time!\par Annie was exploring a nearby ancient temple when she saw something strange. She noticed that someone had reversed the door handle and she couldn't get in! Annie tried her best to open the door but it wouldn't budge.\par Annie's family came and tried to & open & VERB \\
16 & the water.\par Suddenly a frog jumped out and said, "What are you doing?"\par The little squirrel replied, "I'm washing this special nut so that it looks clean and smooth!"\par The frog looked pleased, and said, "That's great! Now it will be even better for you to & wash & VERB \\
17 & jungle with lots of trees and animals. In the jungle lived a leopard. The leopard was very fast and had lots of spots on his fur. He liked to run and play with his friends.\par One day, the leopard hurt his paw while running. He couldn't run or play anymore. His friends wanted to & play & VERB \\
18 & too high.\par "What's wrong?", her mom asked.\par "It's the law," Susan replied. "It won't fly higher than the trees."\par Her mom nodded and smiled. "But you know what? Maybe if you go to the other side of the field, you can & fly & VERB \\
19 & found the chain wrapped tightly around a tree. He sang out as loud as he could, trying to get the tree's attention.\par "I'm stuck!" said the tree. "Can you help me?"\par Sam thought for a moment. He flapped his wings and said, "Yes, I can & help & VERB \\
\hline
\end{longtable}

\subsection{Past tense verb mode in layer 15}

\begin{longtable}{@{}r>{\raggedright\arraybackslash}p{0.62\textwidth}>{\raggedright\arraybackslash}p{0.12\textwidth}l@{}}
\hline
\textbf{index} & \textbf{window} & \textbf{Next TS word} & \textbf{POS} \\
\hline
\endfirsthead
\hline
\textbf{index} & \textbf{window} & \textbf{Next TS word} & \textbf{POS} \\
\hline
\endhead
0 & The coal could not speak, but it started to turn again in the sun. The little boy watched as it spun around and around. His eyes were wide with amazement. He thought it was so cool how the coal moved around in the sun!\par The little boy smiled and said goodbye to the coal. As he & smiled & VERB \\
1 & One day, she found a magic ring in the park. The ring was shiny and had a pretty jewel on it. Lily put the ring on her finger and made a wish. Suddenly, a fairy appeared and said, "Your wish will come true, but you must be serious."\par Lily thought hard about what she & did & VERB \\
2 & girl named Lily. She loved to explore the forest near her house. One day, she went on a walk and discovered a beautiful flower. It was the perfect shade of pink and smelled so nice.\par As she continued walking, she saw smoke in the distance. She followed the smoke and found a big fire. She & followed & VERB \\
3 & . One day mum said, "Daisy, can you manage to bring me a vegetable from the garden?" Daisy smiled and said, "Yes!"\par Daisy went to the garden and looked around. She found tomatoes, peppers, and carrots. She tried to pick a tomato, but it was too heavy. Eventually, she & tried & VERB \\
4 & to play hide and seek with her friends. One day, Lily's friend Max told her about a special hiding spot in the park. Lily was excited to search for it.\par Lily looked everywhere in the park, but she couldn't find the special spot. Just when she was about to give up, she & saw & VERB \\
5 & gifts. Sara smiled and thanked them. They all sat around the table.\par "Are you ready, Sara?" Mommy asked.\par "Yes, Mommy!" Sara said.\par Mommy lit the candles and gave Sara the cake. Sara held the cake carefully. She felt the warmth and the weight of it. She & felt & VERB \\
6 & Lily liked to play with her dad's old things in the attic. She found a big box with a machine that had buttons and a microphone. She wanted to record her voice and hear it again. She pressed a button and said, "Hello, I am Lily. I like to sing and dance." Then she & pressed & VERB \\
7 & wrong?â€\par The voice said â€œIâ€™m a helpless rabbit, and I hopped onto this rock to get away from a hawk. But I canâ€™t get off! Can you help me?â€\par Timmy smiled and said â€œOf course!â€ He carefully & looked & VERB \\
8 & Her friend said, "I lost my toy."\par Lily thought for a moment and then said, "Let's look for it together!" They searched high and low but couldn't find the toy. Suddenly, Lily's friend saw something deep in the bushes. It was the toy! She smiled and & said & VERB \\
9 & down and explained that it was only a bell and it was harmless.\par The fairy thanked the hunter for his kindness. She asked him to take the bell to show others. The hunter was happy to accept. He thanked the fairy and went on his way. The hunter was so excited to find the bell, he & asked & VERB \\
10 & too slow, and she would feel very tired or dizzy. The doctor said she needed an operation to fix it. He said it was a safe and simple operation, but Anna was still scared.\par The day of the operation came. Anna had to wear a special gown and a bracelet with her name on it. She & had & VERB \\
11 & of grass near a lake. The grass was rough, but it felt cozy and safe. He settled down to make the patch his new home.\par Suddenly, the clock started ringing. "Oh no!" Mic thought, "It must be time to go back home!" He hurried away, and as he walked, he & thought & VERB \\
12 & were both very excited and wanted to be the fastest.\par The dog said, "Ready, set, go!"\par The bear replied, "Not yet! I need one more minute!"\par The dog was getting upset and said, "Come on! We must start!"\par But the bear did not listen and just & said & VERB \\
13 & was shining brightly, and they could hear the waves crashing against the shore.\par The father smiled and said, "Let's go swimming!"\par The little boy was so excited that he ran to the water. He was a brave swimmer and paddled out far in to the sea.\par Suddenly he & ran & VERB \\
14 & He had to use all of his strength to protect the town from an incoming storm.\par The giant worked hard and knew what he had to do. He used his giant arms to hold up the storm clouds, and he used his strong legs to keep the ground strong. The people cheered!\par The town never & knew & VERB \\
15 & mom gave her a task to clean her room. Lily didn't want to clean her room but her mom insisted. She started to pick up her toys and put them in the toy box. Suddenly, she found a purple toy that she had lost. She was so happy and showed it to her mom. Her mom & said & VERB \\
16 & climb the hill. They used their hands and feet. They felt the grass tickle their skin. They saw the sky and the trees. They were happy.\par But then, Tom slipped. He lost his balance. He had a fall. He rolled down the hill. He bumped his head and his knee. He & rolled & VERB \\
17 & t hurt you if you don't get too close."\par Tommy felt brave and went closer to the fox. He admired him from a safe distance and watched him mark the woods.\par The fox turned and looked at Tommy. Then he walked away and continued to mark the woods with his scent.\par Tommy smiled and & walked & VERB \\
18 & too high".\par The little girl started to cry. She felt so guilty for wanting the toy.\par Then a fairy godmother appeared. She said, "I will tell you a secret. If you can solve my riddle, I will give you the toy."\par The little girl was very excited. She quickly & said & VERB \\
19 & into small pieces. Sarah watched and smiled as she saw the peppers being snapped.\par John gave Sarah a pepper to try. She popped the pepper into her mouth and tasted the delicious flavor. She said, "Mmm, yummy! These peppers are very tasty!"\par John grabbed some more peppers and smiled. He & said & VERB \\
\hline
\end{longtable}

\subsection{Noun mode in layer 15}
\begin{longtable}{@{}r>{\raggedright\arraybackslash}p{0.62\textwidth}>{\raggedright\arraybackslash}p{0.12\textwidth}l@{}}
\hline
\textbf{index} & \textbf{window} & \textbf{Next TS word} & \textbf{POS} \\
\hline
\endfirsthead
\hline
\textbf{index} & \textbf{window} & \textbf{Next TS word} & \textbf{POS} \\
\hline
\endhead
0 & . Bluey asked, "Why are you crying, little bunny?" The bunny replied, "I have a boo-boo on my knee and it hurts."\par Bluey said, "Don't worry, I can help. Let me wipe your tears and pick you up." Bluey wiped the bunny's & tears & NOUN \\
1 & tree. The tree had many pears. The cat loved to eat the pears.\par One day, the cat was tired. It wanted to rest. The cat found a soft spot under the tree. The cat lay down on the soft spot and closed its eyes.\par While the cat rested, a pear fell from the & tree & NOUN \\
2 & was broken, so every morning it would come back down and try to fix it.\par The square worked and worked, but it soon became tired. Every day it tried to fix the broken houses but nothing it did seemed to help.\par Then one day, something amazing happened--it began to rain. The & rain & NOUN \\
3 & â€™t you try this sport?â€ asked her father. He threw a Frisbee and it went flying in the air.\par Lily ran to catch it. She jumped and caught the Frisbee in her hands.\par â€œWow! That was so much fun!â€ Lily said.\par Her & father & NOUN \\
4 & a kind deer named Dolly appeared and offered to help Benny escape from the forest. Benny was so grateful and thanked Dolly for her rare act of kindness. "I forgive you for getting lost," Dolly said to Benny. "Let's go home now."\par Benny and Dolly made their way back to Benny's & home & NOUN \\
5 & if you are good. But first, eat your nice waffle. It is still warm and yummy."\par Lily nodded. She picked up her fork and took a bite of her nice waffle. It was warm and yummy. She said, "Mmm, this is a good waffle. Thank you, mom."\par Her & mouth & NOUN \\
6 & didn't want to play with her. Mama put her arm around Little Girl and kept her close.\par Little Girl started to cry and Mama asked her why she was so upset. She told Mama that she was sad because she was lonely and the bird didn't want to invite her to play. Mama hugged Little & Girl & PROPN \\
7 & "I'm lost," the puppy said. "I can't find my home."\par Lily and Ben felt sorry for the puppy. They wanted to help.\par "Don't be sad, puppy," Ben said. "We'll help you find your home."\par They crawled out of the & puppy & NOUN \\
8 & a time, there was a little girl named Lily. She loved to play with her toys and draw pictures. One day, she was playing with her dolls when she heard her mom's voice from the kitchen.\par "Lily, can you help me in the kitchen?" her mom asked.\par Lily ran to the & kitchen & NOUN \\
9 & made a bed for the bird and gave it some water. She hoped the bird would heal soon.\par The next day, she checked on the bird. It looked happier and stronger. It flapped its wing and chirped. Lila smiled and opened the box. She carried the bird outside and set it free. The & bird & NOUN \\
10 & to leave. He had to go back to his cave and take a nap. The bear said goodbye to his friends and started to walk away. But then, he remembered the arrow. He didn't want to leave it behind. So, he turned around and went back to get it. The bear picked up the & bear & NOUN \\
11 & wall and it looked very fun.\par But then, Tim's mom came into the room. She saw the paint on the wall and was very mad. She said, "You boys made a big mess! You need to clean it up right now!" Tim and Sam were sad because they had to clean up the & mess & NOUN \\
12 & His name was Thomas. He was three years old and very curious.\par One day, Thomas wanted to explore the world, so he went for a walk. On his walk, he met a bunny. Thomas was so excited to meet the bunny, so he edged closer to give it a hug.\par The & day & NOUN \\
13 & it but it took her a long time and she was delayed in the race.\par Lily was embarrassed that she couldn't keep up with her friends. She tried to run faster but she couldn't catch up. As she was running, she tripped and fell. She scraped her knee and needed a stitch. Her & friends & NOUN \\
14 & feel of it in his hands and looked forward to taking it home.\par Back at home, Mark found the perfect spot for his new candle. He proudly put it in his bedroom and promised to take care of it. Whenever he felt sad or scared, he could look at the candle and remember the fun day at the & end & NOUN \\
15 & walk in the park.\par As he stepped into the park, he saw a big stone sitting by itself. He wanted to see if he could make it move. He got closer and discovered the stone was very fit.\par So the little boy decided to jump on the stone. He leapt up and landed on the & stone & NOUN \\
16 & big tree that was shaking and trembling. Timmy was scared because he didn't know why the tree was shaking.\par As Timmy walked deeper into the forest, he saw a mysterious cave. He was curious and wanted to know what was inside. He went inside the cave and saw a weird creature. The & creature & NOUN \\
17 & was a little train. The train was very slow and it liked to take its time. One day, the train was chugging along when it saw a big pot with steam coming out of it. The train stopped to take a look and saw that there was a man lying on the ground next to the pot. The & train & NOUN \\
18 & tail and a smiley face.\par One day, Ben and Lily want to fly the kite. They go to the park with their mom. The park has a lot of grass and trees. The wind is blowing. Ben and Lily take out the kite. They hold the string. They run and lift the kite. The & kite & NOUN \\
19 & her seat, she saw a rabbit hopping around. This made her smile. She tried to understand what the little animal wanted. She even said 'hello' to it, but it didn't respond.\par Rosa decided to get closer and take a better look. When she got to the edge of the porch, the & porch & NOUN \\
\hline
\end{longtable}

\section{Robustness checks} \label{robust}

\subsection{Robustness check for the modular arithmetic/MLP experiment}

\begin{figure}
\centering
\includegraphics[width=\linewidth]{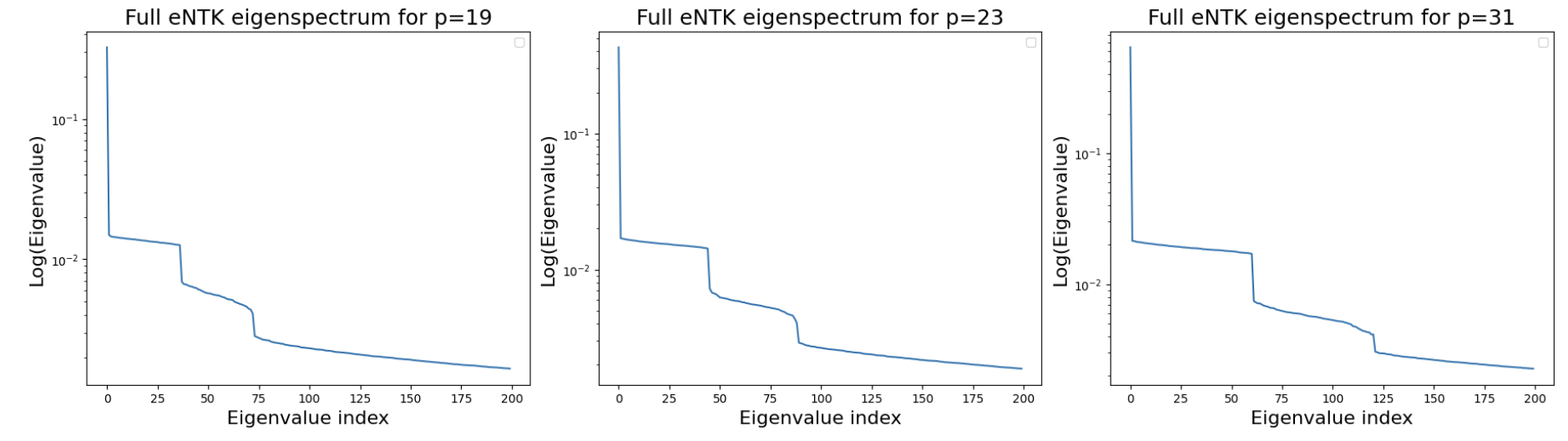}
    \caption{As a robustness check for the modular arithmetic/MLP experiment in section \ref{s3}, we reproduce Fig. 1 for different values of $p$. 
    }
    \label{mlp_robustness}
\end{figure}

In Fig. \ref{mlp_robustness}, we reproduce Fig. \ref{maspectrum} for the MLP trained on modular arithmetic at the different hyperparameters $(p=19, \alpha = 0.8)$, $(p=23,\alpha = 0.75)$, and $(p=31, \alpha=0.65)$. In each case, we find that the eNTK spectrum has two cliffs of size $4\lfloor \frac p 2\rfloor$.

\subsection{Robustness check for the modular arithmetic/Transformer experiment}
In Table \ref{t5}, we reproduce Table \ref{t2} of the Frobenius overlaps between the leading layerwise eNTK eigendirections in the Transformer/modular arithmetic experiment of section \ref{s4} with Fourier families at the (training run-dependent) key frequencies, for the first four random seeds in sequential order besides the one reported in the main text. 

In each case, we train the model to a fixed budget of 4000 epochs and report the number of key frequencies $n$ that appeared in each training run; the test accuracy after 4000 epochs; the squared Frobenius overlap between the first $n+1$ eNTK eigenvectors at each layer and various Fourier families at the key frequencies appearing in each respective training run; and in the case of the MLP$_{out}$ and unembedding layers, the squared Frobenius overlap between the next $n$ eNTK eigenvectors with the Fourier families at the key frequencies appearing in each training run. 

This table shows that the pattern reported in Table \ref{t2} is robust across training runs. 

\begin{table}[t]
  \caption{Squared Frobenius overlap of leading layerwise eNTK eigendirections in the Transformer/modular arithmetic experiment of section \ref{s4} with Fourier families at key frequencies across random seeds. Bolded numbers denote $>$50\% alignment with a particular feature family.  See the text in Appendix \ref{robust} for details on how we define the overlaps in the table.}
  \label{t5}
  \begin{center}
    \begin{small}
      \begin{sc}
        \begin{tabular}{lccccccr}
          \toprule
          Run & No. of key frequencies & Test acc. & Layer  & A+B & sum & diff & ctrl  \\
          \midrule
         {\bf 1} & 4 & 96.9\% &  OV    & {\bf 6.65}/8 & {0.29} & 0.09 & 0.21 \\  
         & & &  MLP$_{in}$   & {\bf 7.34}/8 & 0.06 & 0.04  & 0.24 \\ 
         & & &   MLP$_{out}$ (1)   & {\bf 7.28}/8 & 0.10 & 0.10 & 0.24 \\
         & & &   MLP$_{out}$ (2)   & 0.13 & {\bf 4.35}/8 & {2.43} & 0.27 \\
         & & &   U (1)   & 0.41& {\bf 6.41}/8 & 0.07 & 0.22 \\
         & & &   U (2)   & {\bf 6.90}/8 & {0.28} & 0.16 & 0.21 \\ 
                  {\bf 2} & 4 & 67.0\% &  OV    & {\bf 5.03}/8 & {0.20} & 0.06 & 0.29 \\  
         & & &   MLP$_{in}$   & {\bf 5.45}/8 & 0.21 & 0.03 & 0.14 \\ 
         & & &   MLP$_{out}$ (1)   & {\bf 5.33}/8 & 0.83 & 0.50 & 0.30 \\
         & & &   MLP$_{out}$ (2)   & 0.88 & {3.43} & 0.77 & 0.23 \\
         & & &   U (1)   & 1.63& {\bf 4.36}/8 & 0.14 & 0.26 \\
         & & &  U (2)   & {\bf 4.32}/8 & 0.72 & 0.32 & 0.26 \\ 
                  {\bf 3} & 2 & 100\%  &  OV    & {\bf 3.73}/4 & {0.22} & 0.03 & 0.15 \\  
         & & &   MLP$_{in}$   & {\bf 3.82}/4 & 0.11 & 0.01  & 0.12 \\ 
         & & &  MLP$_{out}$ (1)   & {\bf 2.92}/4 & 1.00 & 0.02 & 0.13 \\
         & & &  MLP$_{out}$ (2)   & 0.98 & {\bf 2.70}/4 & 0.23 & 0.14 \\
         & & & U (1)   & 0.00& {\bf 3.94}/4 & 0.01 & 0.12 \\
         & & &  U (2)   & {\bf 3.94}/4 & 0.00 & 0.02 & 0.15 \\ 
                  {\bf 4} & 3 & 91.9\%  &  OV    & {\bf 4.65}/6 & {0.18} & 0.10 & 0.11 \\  
         & & &  MLP$_{in}$   & {\bf 4.63}/6 & 0.09 & 0.03  & 0.20 \\ 
         & & &  MLP$_{out}$ (1)   & {\bf 3.72}/6 & 2.01 & 0.08 & 0.19 \\
         & & &  MLP$_{out}$ (2)   & 1.00 & {2.30} & 1.59 & 0.20 \\
         & & &  U (1)   & 1.08& {\bf 4.26}/6 & 0.14 & 0.20 \\
         & & &  U (2)   & {\bf 3.62}/6 & 0.12 & 1.42 & 0.21 \\ 
          \bottomrule
        \end{tabular}
      \end{sc}
    \end{small}
  \end{center}
  \vskip -0.1in
\end{table}

\subsection{Robustness check for the experiment with Gemma-3-270M}
In Table \ref{t6}, we reproduce Table \ref{t3} for  random draws of the stratified dataset at the next three random seeds in sequential order besides the one reported in the main text. The dataset is the main source of randomness in this experiment, since for a given dataset we compute (approximate) eNTK eigendirections on a fixed pretrained model, which is deterministic up to numerical Lanczos stability. This table shows that the pattern reported in Table \ref{t3} is robust across dataset draws.

\begin{table}
  \caption{Reproduction of Table \ref{t3} for random draws of the stratified dataset at the next three seeds in sequential order from the one reported in the main text. In Run 3, ``VerbForm=Inf" fell below the $k=30$ example threshold that we used to decide which morphological labels to retain.}
  \label{t6}
  \begin{center}
    \begin{small}
        \begin{tabular}{llcccc}
          \toprule
           Run & Feature & Feature type & Best eNTK AUROC & Best PCA AUROC & best eNTK layer \\
          \midrule
         {\bf 1} &  ADV & pos & {\bf 0.826242} & 0.771484 & 16 \\
         &  CCONJ & pos & {\bf 0.841134} & 0.835240 & 11 \\
         &  DET & pos & {\bf 0.787493} & 0.769845 & 16 \\
         &  NOUN & pos & {\bf 0.969064} & 0.930490 & 16 \\
         &  PRON & pos & {\bf 0.862060} & 0.776472 & 17 \\
         &  PROPN & pos & {\bf 0.935198} & 0.916295 & 12 \\
         & PUNCT & pos & {\bf 0.890137} & 0.889858 & 14 \\
         &  VERB & pos & {\bf 0.928781} & 0.918562 & 12 \\
         &  Number=Plur & morph & {\bf 0.819168} & 0.781181 & 14 \\
         &  Number=Sing & morph & {\bf 0.768686} & 0.715625 & 15 \\
         &  Person=3 & morph & {\bf 0.798208} &  0.762186 & 16 \\
         &  Tense=Past & morph & {\bf 0.980086} & 0.960341 & 16 \\
         & VerbForm=Fin & morph & {\bf 0.978913} & 0.946224 & 13 \\
         & VerbForm=Inf & morph & {\bf 0.981773} & 0.890726 & 15\\
         {\bf 2} &  ADV & pos & {\bf 0.770264} & 0.757638 & 14 \\
         &  CCONJ & pos & {0.831334} & {\bf 0.849330} & 12 \\
         &  DET & pos & {\bf 0.826146} & 0.749302 & 15 \\
         &  NOUN & pos & {\bf 0.965751} & 0.905064 & 16 \\
         &  PRON & pos & {\bf 0.795864} & 0.741037 & 14 \\
         &  PROPN & pos & {\bf 0.937326} & 0.902030 & 12 \\
         & PUNCT & pos & {0.852260} & {\bf 0.861223} & 14 \\
         &  VERB & pos & {\bf 0.976249} & 0.911063 & 16 \\
         &  Number=Plur & morph & {\bf 0.847862} & 0.758050 & 16 \\
         &  Number=Sing & morph & {\bf 0.756683} & 0.730272 & 15 \\
         &  Person=3 & morph & {0.759291} &  {\bf 0.783540} & 17 \\
         &  Tense=Past & morph & {\bf 0.987064} & 0.965023 & 16 \\
         & VerbForm=Fin & morph & {\bf 0.987355} & 0.955621 & 16 \\
         & VerbForm=Inf & morph & {\bf 0.999438} & 0.985718 & 16\\
         {\bf 3} &  ADV & pos & {\bf 0.820328} & 0.774309 & 15 \\
         &  CCONJ & pos & {\bf 0.851911} & {0.829520} & 12 \\
         &  DET & pos & {\bf 0.778948} & 0.758475 & 17 \\
         &  NOUN & pos & {\bf 0.958984} & 0.895612 & 12 \\
         &  PRON & pos & {\bf 0.821254} & 0.785121 & 17 \\
         &  PROPN & pos & {\bf 0.929234} & 0.897670 & 12 \\
         & PUNCT & pos & {0.843054} & {\bf 0.858852} & 16 \\
         &  VERB & pos & {\bf 0.957171} & 0.908343 & 14 \\
         &  Number=Plur & morph & {0.819804} & {\bf 0.838619} & 14 \\
         &  Number=Sing & morph & {\bf 0.816651} & 0.779407 & 16 \\
         &  Person=3 & morph & {\bf 0.826155} &  {0.820603} & 15 \\
         &  Tense=Past & morph & {\bf 0.973199} & 0.946875 & 16 \\
         & VerbForm=Fin & morph & {\bf 0.995686} & 0.953488 & 14 \\
          \bottomrule
        \end{tabular}
    \end{small}
  \end{center}
\end{table}

%
%



\end{document}